\documentclass{article} 
\usepackage{iclr2015,times}
\usepackage{hyperref}
\usepackage{url}
\usepackage{graphicx}
\usepackage{amsmath,amssymb}
\usepackage{array}
\usepackage{multirow}
\usepackage{color,colortbl}
\usepackage[compact]{titlesec}
\usepackage{array}
\usepackage{enumitem}
\usepackage{float}

\usepackage{lineno} 

\newcommand{\dict}{\mathcal{W}}
\newcommand{\chars}{\mathcal{C}}

\definecolor{Gray}{gray}{0.95}

\title{Deep Structured Output Learning for\\ Unconstrained Text Recognition}

\author{
Max Jaderberg\thanks{Current affiliation Google DeepMind. \textsuperscript{+}Current affiliation University of Oxford and Google DeepMind.},~~Karen Simonyan\textsuperscript{*}, Andrea Vedaldi \& Andrew Zisserman\textsuperscript{+}\\
Visual Geometry Group, Department of Engineering Science, University of Oxford\\
\texttt{\{max,karen,vedaldi,az\}@robots.ox.ac.uk} \\ 
}

\def\ie{\emph{i.e.}}

\iclrfinalcopy 

\iclrconference 

\begin{document}

\maketitle

\begin{abstract}
We develop a representation suitable for the unconstrained recognition
of words in natural images, where unconstrained means that there is no
fixed lexicon and words have unknown length.

To this end we propose a convolutional neural network (CNN) based architecture which incorporates a Conditional Random Field (CRF) graphical model, taking the whole word image as a single input. The unaries of the CRF are provided by a CNN that predicts characters at each position of the output, while higher order terms are provided by another CNN that detects the presence of N-grams. We show that this entire model (CRF, character predictor, N-gram predictor) can be jointly optimised by back-propagating the structured output loss, essentially requiring the system to perform multi-task learning, and training requires only synthetically generated data. The resulting model is a more accurate system on standard real-world text recognition benchmarks than character prediction alone, setting a benchmark for systems that have not been trained on a particular lexicon. In addition, our model achieves state-of-the-art accuracy in lexicon-constrained scenarios, without being specifically modelled for constrained recognition. To test the generalisation of our model, we also perform experiments with random alpha-numeric strings to evaluate the method when no visual language model is applicable.

\end{abstract}

\section{Introduction}

In this work we tackle the problem of \emph{unconstrained text
recognition} -- recognising text in natural images without restricting
the words to a fixed lexicon or dictionary. 
Usually this problem is decomposed into a word detection
stage followed by a word recognition stage. The word detection stage
generates bounding boxes around words in an image, while the
word recognition stage takes the content of these bounding boxes and recognises the text within. 

This paper focuses on the text recognition stage, developing a model based on deep convolutional neural networks (CNNs)
(\cite{Lecun98}).
Previous methods using CNNs for word recognition (discussed in more detail in section Section~\ref{sec:related}) has
either constrained (\cite{Jaderberg14d}) or heavily weighted
(\cite{Bissacco13}) the recognition results to be from a dictionary of
known words. This works very well when training and testing are limited to a fixed
vocabulary, but does not generalise to where previously unseen or
non-language based text must be recognised -- for example for generic
alpha-numeric strings such as number plates or phone numbers.

The shift of focus towards a model which performs accurately without a
fixed dictionary increases the complexity of the text recognition
problem. To solve this,  we propose a novel CNN
architecture (Figure~\ref{fig:path}) employing a 
\emph{Conditional Random Field} (CRF) whose unary terms are 
outputs of a CNN character predictor, which are position-dependent, 
and whose higher order terms are 
outputs of a CNN N-gram predictor, which are position-independent. 
The recognition result is then obtained by
finding the character sequence that maximises the CRF score, enforcing
the consistency of the individual predictions.

The CRF model builds on our previous work where
we explored 
dictionary-based recognition (\cite{Jaderberg14c}) for two
scenarios: the first was to train a different
CNN character classifier for each position in the word being recognised,
using the whole image of the word as input to each classifier
(an idea also expored by~\cite{Goodfellow13}); the
second was to construct a CNN predictor to detect the N-grams
contained in the word, effectively encoding the text as a
bag-of-N-grams. 

The dictionary-free joint model proposed here is
trained by defining a structured output learning problem, and
back-propagating the corresponding \emph{structured output loss}. This
formulation results in multi-task learning of both the character
and N-gram predictors, and additionally learns how to
combine their representations in the CRF, resulting in
more accurate text recognition.

The result is a highly flexible
text recognition system that achieves excellent unconstrained
text recognition performance as well as state-of-the-art recognition
performance when using standard dictionary constraints. While performance is measured on real images as contained in standard text recognition benchmarks, all results are obtained by training the model~\emph{purely on synthetic data}. The model is evaluated on this synthetic data as well in order to study its performance under different scenarios.


Section~\ref{sec:related} outlines work related to ours. Section~\ref{sec:characters} reviews the character sequence model and Section~\ref{sec:ngrams} the
bag-of-N-grams model. Section~\ref{sec:joint} shows how these predictors can be combined to form a joint CRF model and formulates the training of the latter as structured-output learning. 
Section~\ref{sec:eval} evaluates these models extensively 
and Section~\ref{sec:conclusion} summarises our findings.

\section{Related Work}
\label{sec:related}

We concentrate here on text recognition methods, recognising from a cropped image of a single word, rather than the text detection stages of scene text
recognition (`text spotting') that generate the word detections.
	
Traditional text recognition methods are based on sequential character
classification, finding characters by sliding
window methods (\cite{Wang11,Wang12,Jaderberg14a}, after which a word
prediction is made by integrating character classifier predictions in a
left-to-right manner. The character classifiers include random
ferns (\cite{Ozuysal07}) in \cite{Wang11}, and CNNs in~\cite{Wang12,Jaderberg14a}. Both~\cite{Wang11} and~\cite{Wang12} use a
small fixed lexicon as a language model to constrain word recognition.

More recent works such as~\cite{Bissacco13,Alsharif13} make
use of over-segmentation methods, guided by a supervised classifier, to
generate candidate character proposals in a single-word image, which are subsequently classified as
true or false positives. For example, PhotoOCR~(\cite{Bissacco13}) uses binarization and a sliding window classifier to generate candidate character regions, with words recognised through a beam search driven by classifier scores and static N-gram language model, followed by a re-ranking using a dictionary of 100k words.~\cite{Jaderberg14a} uses the convolutional nature of CNNs to generate response maps for characters and bigrams which are integrated to score lexicon words.


In contrast to these approaches based on character
classification, the work
by~\cite{Almazan14,Gordo14,Goel13,Rodriguez13,Novikova12,Mishra12} instead uses the
notion of holistic word recognition.~\cite{Mishra12,Novikova12} still
rely on explicit character classifiers, but construct a graph to infer
the word, pooling together the full word
evidence. \cite{Rodriguez13} use aggregated Fisher
Vectors~(\cite{Perronnin10}) and a Structured SVM
framework to create a joint word-image and text
embedding. \cite{Almazan14} and more recently \cite{Gordo14} also formluate joint embedding spaces, achieving impressive results with minimal training data.~\cite{Goel13} use whole word-image features to recognize words by comparing to simple black-and-white font-renderings of
lexicon words. In our own previous work (\cite{Jaderberg14c,Jaderberg14d}) we use large CNNs acting on the full word image region to perform 90k-way classification to a dictionary word.

It should be noted that all the methods make use of strong static
language models, either relying on a constrained dictionary or
re-ranking mechanism. 

\cite{Goodfellow13} had great success using a CNN with multiple position-sensitive character classifier outputs (closely related to the character sequence model in Section \ref{sec:characters}) to perform street number recognition. This model was extended to CAPTCHA sequences (up to 8 characters long) where they demonstrated impressive performance using synthetic training data for a synthetic problem (where the generative model is known), but we show that synthetic training data can be used for a real-world data problem (where the generative model is unknown).

There have been previous uses of graphical models with back-propagated loss functions for neural networks, such as the early text recognition work of \cite{Lecun98} to combine character classifier results on image segmentations. Another example is the recent work of \cite{Tompson14} for human pose estimation, where an MRF-like model over the distribution of spatial locations for each body part is constructed, incorporating a single round of message-passing.

\section{CNN Text Recognition Models}
We now review the component CNN models, originally presented in our tech report \cite{Jaderberg14c}, 
that form the basis of our joint model in Section~\ref{sec:joint}.

\subsection{Character Sequence Model Review}
\label{sec:characters}

\begin{figure}
\begin{center}
\begin{tabular}{cc}
\includegraphics[width=0.5\linewidth]{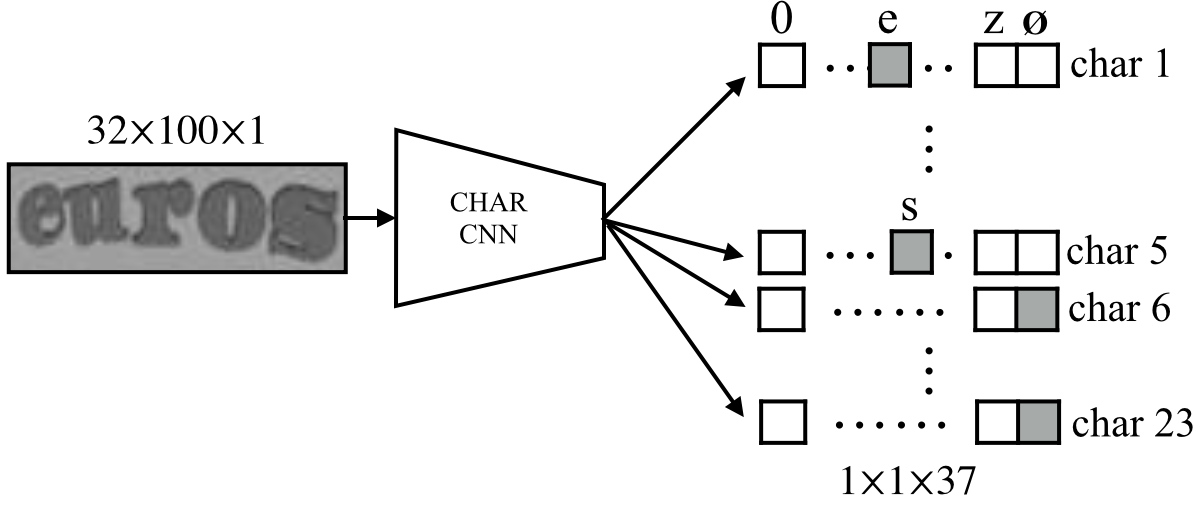}&
\includegraphics[width=0.36\linewidth]{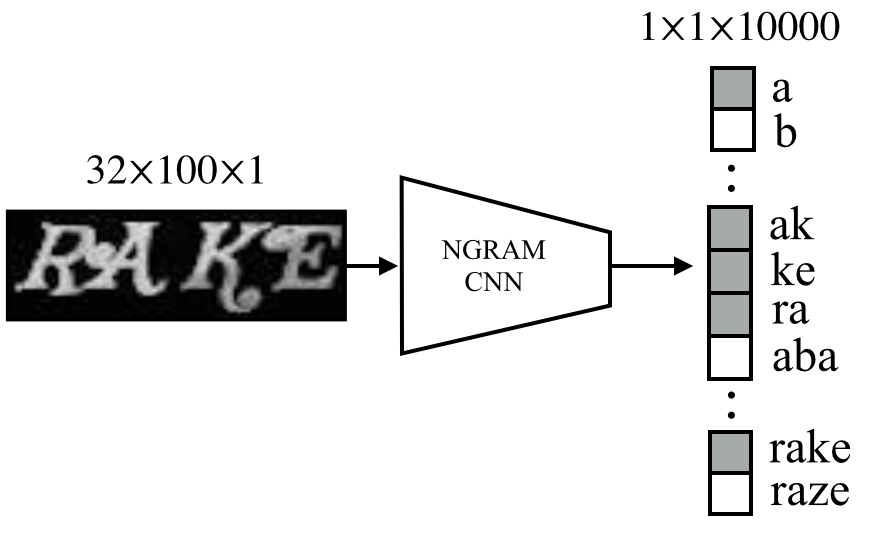}\\
(a) & (b)
\end{tabular}
\vspace{-1em}
\caption{\small (a) The character sequence model. A word image is recognised by predicting the character at each position in the output, spelling out the text character by character. Each positional classifier is learnt independently but shares a jointly optimised set of features. (b) The N-gram encoding model. The recognised text is represented by its bag-of-N-grams. This can be thought of as 10k independently trained binary classifiers using a shared set of jointly learnt features, trained to detect the presence of a particular N-gram.}
\label{fig:basenets}
\end{center}
\end{figure}

In this section we describe our character sequence model. This model
encodes the character at each position in the word and so predicts the
sequence of characters in an image region (hereafter we simply refer to the
image region as an image). Each position in the word is
modelled by an independent classifier acting on a shared set of
features from a single CNN. By construction, this model makes no
assumptions about the underlying language and allows completely
unconstrained recognition.

A word $w$ of length $N$ is modelled as a sequence of characters such that $w = (c_1, c_2, \dots, c_N)$ where each $c_i \in \chars = \{1,2,\dots,36\}$ represents a character at position $i$ in the word, from the set of 10 digits and 26 letters. Each $c_i$ can be predicted with a single classifier, one for each character in the word. However, since words have variable length $N$ which is unknown at test time, we fix the number of characters to $N_\text{max}$ (here set to 23), the maximum length of a word in the training set, and introduce a null character class. Therefore a word is represented by a string $w \in (\chars \cup \{\phi\})^{N_\text{max}}$. 

For a given input image $x$, we want to return the estimated word $w^*$ which maximises $P(w^*|x)$. Since we seek an unconstrained recognition system with this model, we assume independence between characters leading to 
\begin{equation}
w^* = \arg\max_{w} P(w|x) = \arg\max_{c_1,c_2,\dots,c_{N_\text{max}}} \prod_{i=1}^{N_\text{max}} P(c_i|\Phi(x)) 
\label{eqn:wordprob}
\end{equation}
where $P(c_i|\Phi(x))$ is given by the classifier 
for the $i$-th position
acting on a single set of shared CNN features $\Phi(x)$. 
The word $w^*$ 
can be computed by taking the most probable character at each position 
$
c_i^* = \arg\max_{c_i \in \chars \cup \{\phi\}} P(c_i|\Phi(x)).
$

The CNN (Figure~\ref{fig:basenets}~(a)) takes the whole word image $x$ as input. Word images can be of different sizes, in particular due to the variable number of characters in the image. However, our CNN requires a fixed size input for all input images. This problem is overcome by simply resampling the original word image to a canonical height and width, without regard to preserving the aspect ratio, producing a fixed size input $x$.

The base CNN has a number of convolutional layers followed by a series of fully connected layers, giving $\Phi(x)$. The full details of the network architecture are given in Section~\ref{sec:implementation}. $\Phi(x)$ is fed to $N_\text{max}$ separate fully connected layers with 37 neurons each, one for each character class including the null character. These fully connected layers are independently softmax normalised and can be interpreted as the probabilities $P(c_i|\Phi(x))$ of the width-resized input image $x$.

The CNN is trained with multinomial logistic regression loss, back-propagation, and stochastic gradient descent (SGD) with dropout regularisation similar to \cite{Hinton12}.



\subsection{Bag-of-N-grams Model Review}
\label{sec:ngrams}

This section describes our second word recognition model, which exploits compositionality to represent words. In contrast to the sequential character encoding of Section~\ref{sec:characters}, words can be seen as a composition of an unordered set of character N-grams, a \emph{bag-of-N-grams}. In the following, if $s\in \chars^N$ and $w\in\chars^M$ are two strings, the symbol $s \subset w$ indicates that $s$ is a substring
of $w$. An $N$-gram
of word $w$ is a substring $s \subset w$ of length $|s|=N$. We will
denote with $G_N(w) = \{ s : s\subset w \,\wedge\,|s|\leq N\}$ the set
of all N-grams of word $w$ of length up to $N$ and with $G_N
=\cup_{w\in\dict} G_N(w)$ the set of all such grams in the
language. For example,
$G_3(\texttt{spires})=\{\texttt{s},\texttt{p},\texttt{i},\texttt{r},\texttt{e},\texttt{sp},\texttt{pi},\texttt{ir},\texttt{re},\texttt{es},\texttt{spi},\texttt{pir},\texttt{ire},\texttt{res}\}$. This method of encoding variable length sequences is similar to the \emph{Wickelphone} phoneme-encoding methods (\cite{Wickelgran69}).

Even for small values of $N$, $G_N(w)$ encodes each word $w\in\mathcal{W}$ nearly uniquely. For example, with $N=4$, this map has only 7 collisions out of a dictionary of 90k words. The encoding $G_N(w)$ can be represented as a $|G_N|$-dimensional binary vector of N-gram occurrences. This vector is very sparse, as on average $|G_N(w)|\approx 22$ whereas $|G_N|=10k$. 

Using a CNN we can predict $G_N(w)$ for a word $w$ depicted in the input image $x$. We can use the same architecture as in Section~\ref{sec:characters}, but now have a final fully connected layer with $G_N$ neurons to represent the encoding vector. The scores from the fully connected layer can be interpreted as probabilities of an N-gram being present in the image by applying the logistic function to each neuron. The CNN is therefore learning to recognise the presence of each N-gram somewhere within the input image, so is an N-gram detector.

With the applied logistic function, the training problem becomes that of $|G_N|$ separate binary classification tasks, and so we back-propagate the logistic regression loss with respect to each N-gram class independently. To jointly train a whole range of N-grams, some of which occur very frequently and some barely at all, we have to scale the gradients for each N-gram class by the inverse frequency of their appearance in the training word corpus. We also experimented with hinge loss and simple regression to train but found frequency weighted binary logistic regression was superior. As with the other model, we use dropout and SGD.

In this model we exploit the statistics of our underlying language in choosing a subset of $|G_N|$ N-grams from the space of all possible N-grams to be modelled. This can be seen as using a language model to compress the representation space of the encoding, but is not restraining the predictive capability for unconstrained recognition. While the encoding $G_N(w)$ is almost always unique for words from natural language, non-language words often contain much fewer N-grams from the modelled set $G_N$ leading to more ambiguous and non-unique encodings. 

\begin{figure}
\begin{center}
\begin{tabular}{cc}
\includegraphics[width=\linewidth]{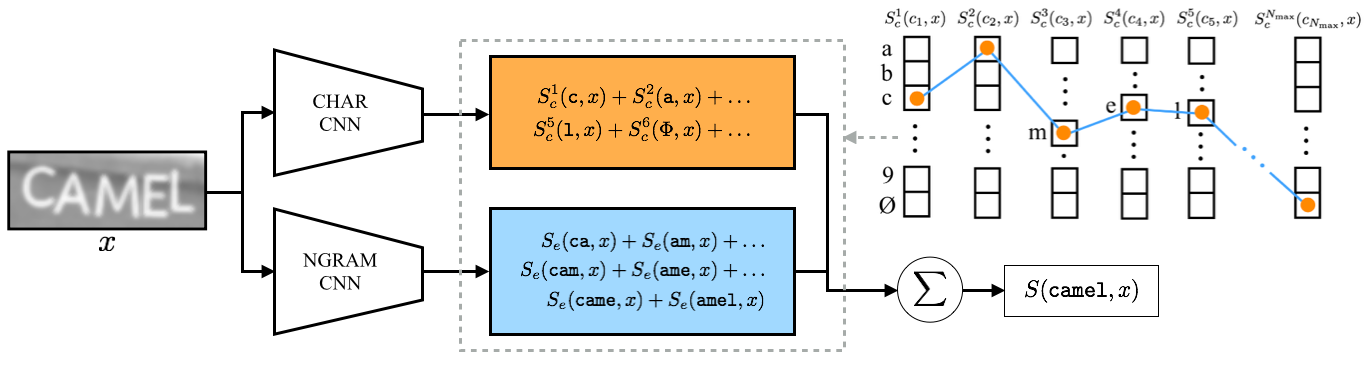}&
\end{tabular}
\vspace{-1.5em}
\caption{\small An illustration of the construction of the path score $S(\texttt{camel},x)$ for the word $\texttt{camel}$. The unary and edge terms used for the score are selected by the path through the graph of character positions shown in the upper right corner. The values of these terms, $S_c(c_i,x)$ and $S_e(s,x)$, where $s \subset w$, are given by the outputs of the character sequence model CNN (CHAR CNN) and the N-gram encoding CNN (NGRAM CNN).}
\label{fig:path}
\end{center}
\end{figure}

\nopagebreak
\section{Joint Model}
\label{sec:joint}

In Section~\ref{sec:characters}, maximising the posterior probability of a character sequence \eqref{eqn:wordprob} is equivalent to maximising the log-score
$
\log P(w|x) = S(w,x) = \sum_{i=1}^{N_\text{max}} S_c^i(c_i,x)\label{eqn:wordscore}
$
where $S_c^i(c_i,x)=\log P(c_i|\Phi(x))$ is the logarithm of the posterior probability of the character at position $i$ in the sequence. The graph associated with this function is a set of nodes, one for each unary term $S_c^i(c_i,x)$, and does not contain any edges. Hence maximising the function reduces to maximising each term individually.


The model can now be extended to incorporate the N-gram predictors of Section~\ref{sec:ngrams}, encoding the presence of N-grams in the word image $x$. The N-gram scoring function $S_e(s,x)$ assigns a score to each string $s$ of length $|s| \leq N$, where $N$ is the maximum order of N-gram modelled. Note that, differently from the functions $S_c^i$ defined before, the function $S_e$ is position-independent. However, it is applied repeatedly at each position $i$ in the word:
\begin{equation}
S(w,x) = 
\sum_{i=1}^{N_\text{max}} S_c^i(c_i,x) 
+ \sum_{i=1}^{|w|} \sum_{n=1}^{\min(N,|w|-i+1)}
S_e(c_ic_{i+1}\dots c_{i+n-1},x).
\label{eqn:jointscore}
\end{equation}
As illustrated in Figure~\ref{fig:path}, the scores $S_c^i(c_i,x)$ are obtained from the CNN character predictors of Section~\ref{sec:characters} whereas the score $S_e(s,x)$ is obtained from the CNN N-gram predictor of Section~\ref{sec:ngrams}; note that the N-gram scoring function is only defined for the subset $G_N$ of N-grams modelled in the CNN; if $s\not\in G_N$, the score $S_e(s,x)=0$ is defined to be zero.

The graph associated with the function~\eqref{eqn:jointscore} has cliques of order $N$; hence, when $N$ is even moderately large, we resort to beam search (\cite{Russel94}) to maximise~\eqref{eqn:jointscore} and find the predicted word $w^*$. 
Also, the score~\eqref{eqn:jointscore} can be interpreted as a potential function defining a word posterior probability as before; however, evaluating this probability would require computing a normalisation factor, which is non-trivial. Instead, the function is trained discriminatively, as explained in the next section.

\paragraph{Structured Output Loss.}

The unary and edge score functions $S_c^i(c_i,x)$ and $S_e(s,x)$, should incorporate the outputs of the character sequence model and N-gram encoding model respectively. A simple way to do this is to apply a weighting to the output of the CNNs after removing the softmax normalisation and the logistic loss:
\begin{equation}
S(w,x) = 
\sum_{i=1}^{N_\text{max}} \alpha_{c_i}^i f_{c_i}^i(x) 
+ \sum_{i=1}^{|w|} \sum_{n=1}^{\min(N,|w|-i+1)}
\beta_{c_ic_{i+1}\dots c_{i+n-1}} g_{c_ic_{i+1}\dots c_{i+n-1}}(x),
\label{eqn:weighted}
\end{equation}
where $f_{c_i}^i(x)$ is the output of the character sequence CNN for character $c_i$ at position $i$ and $g_{s}(x)$ is the output of the N-gram encoding CNN for the N-gram $s$. If desired, the character weights $\alpha = \{\alpha_{c_i}^i\}$ and edge weights $\beta = \{\beta_s\}$ can be constrained to be shared across different characters, character positions, different N-grams of the same order, or across all N-grams.

The sets of weights $\alpha$ and $\beta$ in Equation~\ref{eqn:weighted}, or any weight-constrained variant of Equation~\ref{eqn:weighted}, can be learnt in a structured output learning framework, encouraging the score of the ground-truth word $w_{\text{gt}}$ to be greater than or equal to the highest scoring \emph{incorrect} word prediction plus a margin, \ie~$S(w_{\text{gt}},x) \geq \mu + S(w^*,x)$ where $S(w^*,x) = \max_{w\not= w_{\text{gt}}} S(w,x)$. Enforcing this as a soft-constraint results in the  convex loss
\begin{equation}
L(x_i,w_{\text{gt},i},S)
=\max_{w\not= w_{\text{gt},i}} \max(0, \mu + S(w,x) - S(w_{\text{gt},i},x_i))
\end{equation}
and averaging over $M$ example pairs $(x_i,w_{\text{gt},i})$ results in the regularised empirical risk objective
\begin{equation}
 E(S) = 
 \frac{\lambda_\alpha}{2}\|\alpha\|^2 + 
 \frac{\lambda_\beta }{2}\|\beta\|^2 + 
 \frac{1}{M}\sum_{i=1}^M 
 L(x_i,w_{\text{gt},i},S).
 \label{eqn:cost}
\end{equation}

However, in the general scenario of Equation~\ref{eqn:weighted}, the weights can be incorporated into the CNN functions $f$ and $g$, resulting in the score
\begin{equation}
S(w,x) = 
\sum_{i=1}^{N_\text{max}} f_{c_i}^i(x) 
+ \sum_{i=1}^{|w|} \sum_{n=1}^{\min(N,|w|-i+1)}
 g_{c_ic_{i+1}\dots c_{i+n-1}}(x),
\label{eqn:cnnscore}
\end{equation}
The functions $f$ and $g$ are defined by CNNs and so we can optimise the parameters of them to reduce the cost in Equation~\ref{eqn:cost}. This can be done through standard back-propagation and SGD. Differentiating the loss $L$ with respect to $S$ gives
\begin{equation}
\frac{\partial L(x,w_{\text{gt}},S)}{\partial S(w^*,x)} = 
\begin{cases}
		1  & \mbox{if } z > 0 \\
		0  & \mbox{otherwise}
\end{cases}
\qquad
\frac{\partial L(x,w_{\text{gt}},S)}{\partial S(w_{\text{gt}},x)} = 
\begin{cases}
		-1  & \mbox{if } z > 0 \\
		0  & \mbox{otherwise}
\end{cases}
\end{equation}
where $z=\max_{w\not= w_{\text{gt},i}} \mu + S(w,x) - S(w_{\text{gt}},x)$. Differentiating the score function of Equation~\ref{eqn:cnnscore} with respect to the character sequence model and N-gram encoding model outputs $f_{c_i}^i$ and $g_s$ gives
\begin{equation}
\frac{\partial S(w,x)}{\partial f_{c}^i} = 
\begin{cases}
		1  & \mbox{if } c_i = c \\
		0  & \mbox{otherwise}
\end{cases},
\qquad
\frac{\partial S(w,x)}{\partial g_{s}} = 
\sum_{i=1}^{|w|-|s|+1}
\mathbf{1}_{\{ c_ic_{i+1}\dots c_{i+|s|-1} = s\}}
\end{equation}
This allows errors to be back-propagated to the entire network. Intuitively, the errors are back-propagated through the CNN outputs which are responsible for margin violations, since they contributed to form an incorrect score.

\begin{figure}
\begin{center}
\begin{tabular}{cc}
\includegraphics[width=0.8\linewidth]{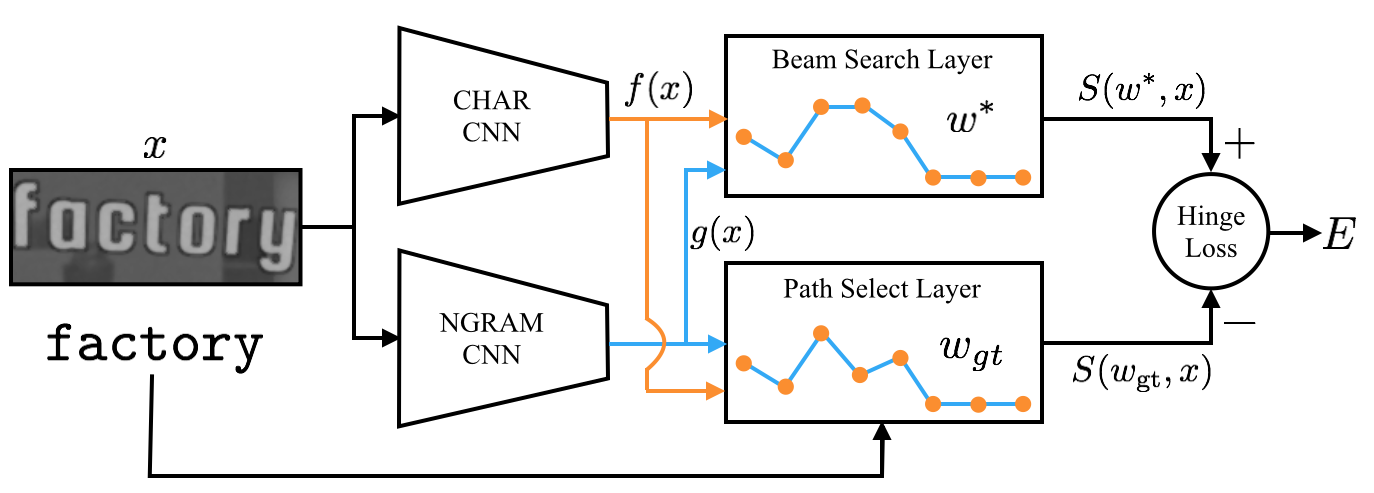}&
\end{tabular}
\caption{\small The architecture for training the joint model, comprising of the character sequence model (CHAR) and and the N-gram encoding model (NGRAM) with structured output loss. The Path Select Layer generates the score $S(w_{\text{gt}},x)$ by summing the inputs of the groundtruth word. The Beam Search Layer uses beam search to try to select the path with the largest score $S(w^*,x)$ from the inputs. The hinge loss implements a ranking loss, constraining the highest scoring path to be the groundtruth path, and can be back-propagated through the entire network to jointly learn all the parameters.}
\label{fig:training}
\end{center}
\end{figure}

Using this structured output loss allows the parameters of the entire model to be jointly optimised within the structure imposed by Equation~\ref{eqn:cnnscore}. Figure~\ref{fig:training} shows the training architecture used. Due to the presence of high order scores in Equation~\ref{eqn:cnnscore}, it is too expensive to exhaustively search the space of all possible paths to find $w^*$, even with dynamic programming, so instead we use beam search to find the approximate highest scoring path.

The structured output loss described in this section bares resemblance to the discriminative Viterbi training introduced by \cite{Lecun98}. However, our model includes higher-order terms, terms of a different nature (N-grams), and uses a structured-output formulation. Furthermore, our method incorporates only a very weak language model, limited to assigning a score of 0 to all N-grams outside a target set $G_N$. Note that this does not mean that these N-grams cannot be recognised (this would require assigning to them a score of $-\infty$); instead, it is a smoothing technique that assigns a nominal score to infrequent N-grams.




\section{Evaluation}
\label{sec:eval}
In this section we evaluate the three models introduced in the previous sections. The datasets used for training and testing are described in Section~\ref{sec:datasets}, the implementation details given in Section~\ref{sec:implementation}, and the results of experiments reported in Section~\ref{sec:experiments}.

\subsection{Datasets}
\label{sec:datasets}
We evaluate our models on a number of standard datasets -- ICDAR 2003, ICDAR 2013, Street View Text, and IIIT5k, whereas for training, as well as testing across a larger vocabulary, we turn to the synthetic Synth90k and SynthRand datasets.

{\bf ICDAR 2003}~(\cite{ICDAR03}) is a scene text recognition dataset,
with the test set containing 251 full scene images and 860 groundtruth
cropped images of the words contained with the full images. We follow
the standard evaluation protocol defined by~\cite{Wang11} and perform
recognition on the words containing only alphanumeric characters
and at least three characters. The test set of 860 cropped word images
is referred to as IC03. The lexicon of all test words is IC03-Full (563 words),
and the per-image 50 word lexicons defined by~\cite{Wang11} and used
in a number of works (\cite{Wang11,Wang12,Alsharif13}) are referred to
as IC03-50. 

{\bf ICDAR~2013} (\cite{ICDAR2013}) test dataset contains 1015 groundtruth cropped word images from scene text. Much of the data is inherited from the ICDAR 2003 datasets. We refer to the 1015 groundtruth cropped words as IC13. 

{\bf Street View Text} (\cite{Wang11}) is a more challenging scene text dataset than the ICDAR datasets. It contains 250 full scene test images downloaded from Google Street View. The test set of 647 groundtruth cropped word images is referred to as SVT. The lexicon of all test words is SVT-Full (4282 words), and the smaller per-image 50 word lexicons defined by~\cite{Wang11} and used in previous works (\cite{Wang11,Wang12,Alsharif13,Bissacco13}) are referred to as SVT-50. 

{\bf IIIT 5k-word} (\cite{Mishra12}) test dataset contains 3000 cropped word images of scene text downloaded from Google image search. Each image has an associated 50 word lexicon (IIIT5k-50) and 1k word lexicon (IIIT5k-1k).

{\bf Synth90k\footnote{\url{http://www.robots.ox.ac.uk/~vgg/data/text/}}} (\cite{Jaderberg14c,Jaderberg14d}) is a dataset of 9 million cropped word images that have been synthetically generated. The synthetic data is highly realistic and can be used to train on and as a challenging test benchmark. The dataset covers 90k different English words, and there are predefined training and test splits with approximately 8 million training images and 900k test images. In addition, we use the same synthetic text engine from \cite{Jaderberg14c,Jaderberg14d} to generate word images with completely random strings of up to 10 uniformly sampled alphanumeric characters. We refer to this dataset as {\bf SynthRand}. The training set consists of 8 million training images and the test set of 900k images. In this corpus there are very few word repetitions (in addition to the random rendering variations). There is a wide range of difficulty in this dataset, from perfectly readable text to almost impossible to read samples.


\subsection{Implementation Details}
\label{sec:implementation}

In the following, the character sequence model is referred to as CHAR, the N-gram encoding model as NGRAM, and the joint model as JOINT. 

The CHAR and NGRAM models both have the same base CNN architecture. The base CNN has five convolutional layers and two fully connected layers. The input is a $32 \times 100$ greyscale image obtained by resizing the word image  (ignoring its aspect ratio) and then subtracting its mean and dividing by its standard deviation. Rectified linear units are used throughout after each weight layer except for the last one. In forward order, the convolutional layers have 64, 128, 256, 512, and 512 square filters with an edge size of 5, 5, 3, 3, and 3. Convolutions are performed with stride 1 and there is input feature map padding to preserve spatial dimensionality. $2 \times 2$ max-pooling follows the first, second and third convolutional layers. The fully connected layers have 4096 units. 

On top of this base CNN, the CHAR model has 23 independent fully connected layers with 37 units, allowing recognition of words of up to $N_\text{max}=23$ characters long. The NGRAM model operates on a selection of 10k frequent N-grams of order $N\leq 4$ (identified as the ones that occur at least 10 times in the Synth90k word corpus, resulting in 36 1-grams, 522 2-grams, 3965 3-grams, and 5477 4-grams). This requires a final fully connected layer on top of the base CNN with 10k units. Therefore, the graph of function~\eqref{eqn:cnnscore} has cliques of sizes at most 4. Beam search uses a width of 5 during training and of 10 during testing. If a lexicon is used to constrain the output, instead of performing beam search, the paths associated with the lexicon words are scored with Equation~\ref{eqn:cnnscore}, and the word with the maximum score is selected as the final result.

The three models are all trained with SGD and dropout regularisation. The learning rates are dynamically decreased as training progresses. The JOINT model is initialised with the pre-trained CHAR and NGRAM network weights and the convolutional layers' weights are frozen during training.

\subsection{Experiments}
\label{sec:experiments}
We evaluate our models on a combination of real-world test data and synthetic data to highlight different operating characteristics.

\paragraph{N-gram Encoding Results.}
The NGRAM model predicts the N-grams contained in input word image. Due to the highly unbalanced nature of this problem (where only 10-20 N-grams are contained in any given image), results are reported as the maximum achieved F-score, computed as the harmonic mean of precision and recall. The latter are computed by sweeping the threshold probability for an N-gram to be classified as present in the word. The maximum achieved F-score on Synth90k is 87.0\% and on IC03 is 87.1\%. This demonstrates that, while not perfect, the NGRAM model accurately models the presence of N-grams in word images.


\setlength{\tabcolsep}{3pt}
\begin{table}[t]
\begin{center}
\begin{tabular}[t]{|l|l||c|c|} 
\hline
\multicolumn{1}{|c|}{\centering Training Data} & 
\multicolumn{1}{c||}{\centering Test Data} &
\multicolumn{1}{c|}{\centering CHAR} &
\multicolumn{1}{c|}{\centering JOINT} \\
\hline\hline
\multirow{4}{*}{Synth90k-train} & Synth90k-test & 87.3 & \bf 91.0\\
 & IC03 & 85.9 & \bf 89.6 \\
  & SVT & 68.0 & \bf 71.7 \\
  & IC13 & 79.5 & \bf 81.8 \\
\hline
Synth1-72k & Synth72k-90k & 82.4 & \bf 89.7  \\
\hline
Synth1-45k & Synth45k-90k & 80.3 & \bf 89.1 \\
\hline
SynthRand & SynthRand & \bf 80.7 & 79.5 \\
\hline
\end{tabular}
\qquad\quad
\vtop{\vspace{-0.8em}\hbox{\includegraphics[width=0.38\textwidth]{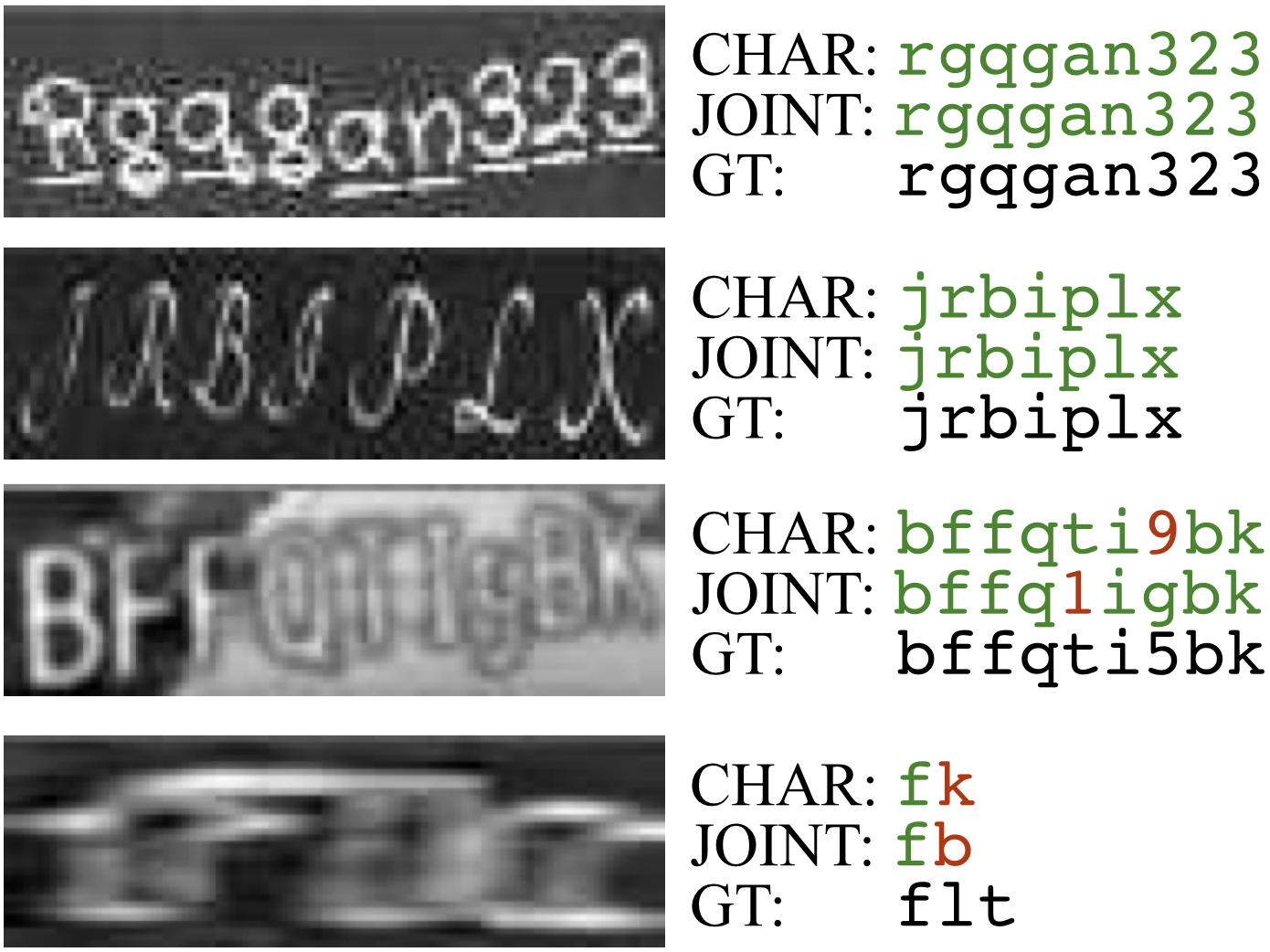}}}%
\end{center}
\vspace*{-1em}
\caption{\small \emph{Left:}  The accuracy (\%) of the character sequence model, CHAR, and the joint model, JOINT. Different combinations of training and test data are evaluated. Synth$x$-$y$ refers to a subset of the Synth90k that only contains words in the label interval $[x,y]$ (word label indices are in random, non-alphabetical order). Training and testing on completely distinct words demonstrates the power of a general, unconstrained recognition model. \emph{Right:} Some results of the CHAR model on the SynthRand test dataset. Letters in red have been predicted incorrectly with the groundtruth (GT) shown below. Notice the range in difficulty of the SynthRand data.}
\label{table:internal}
\end{table}

\paragraph{Character Sequence and Joint Model Results.}

\begin{figure}
\begin{center}
\begin{tabular}{ccc}
\includegraphics[width=0.3\linewidth]{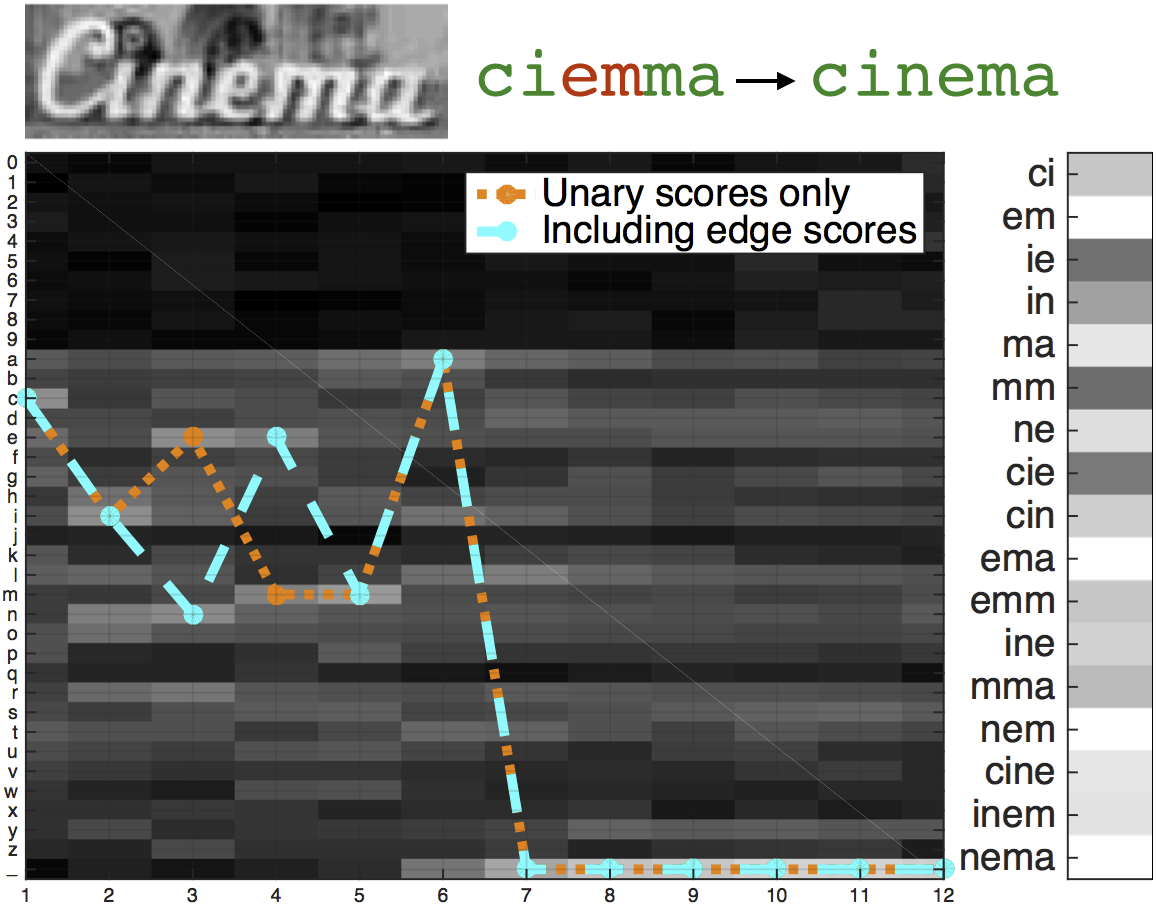}&
\includegraphics[width=0.3\linewidth]{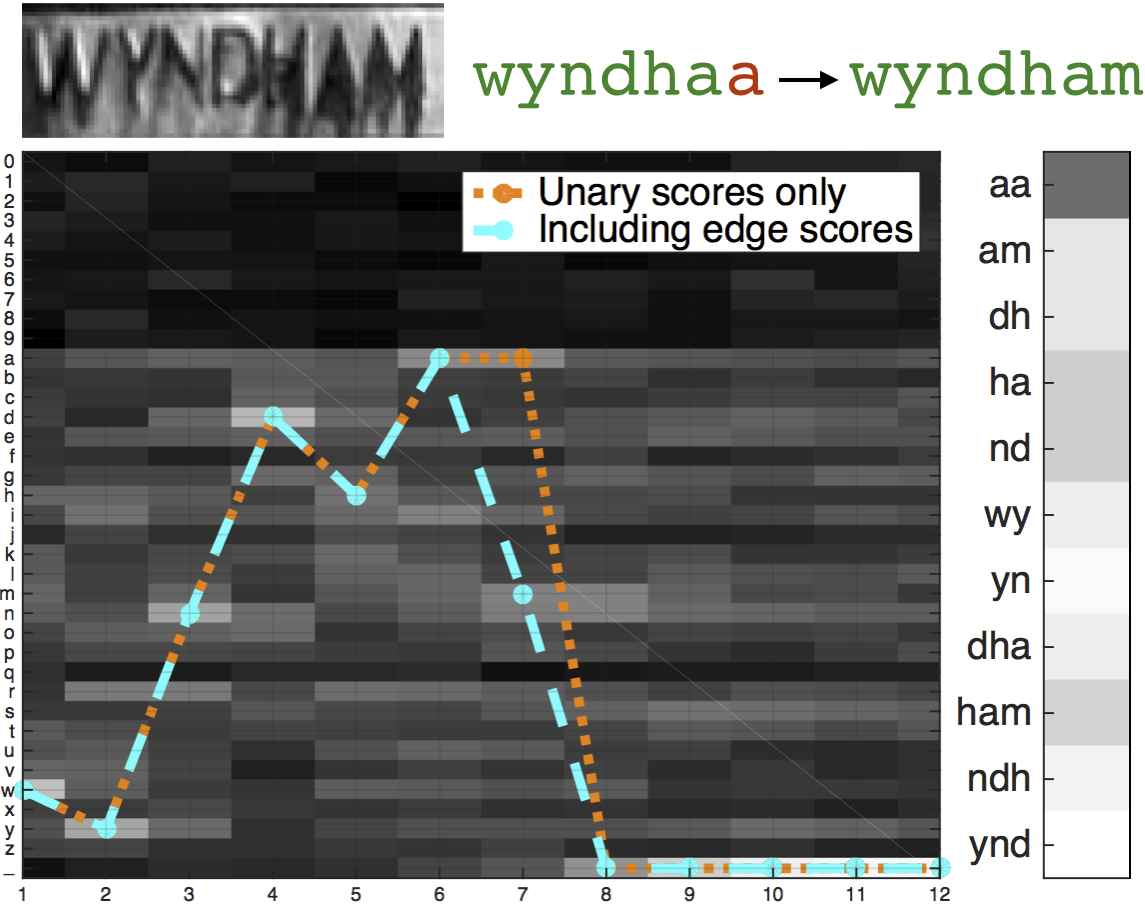}&
\includegraphics[width=0.32\linewidth]{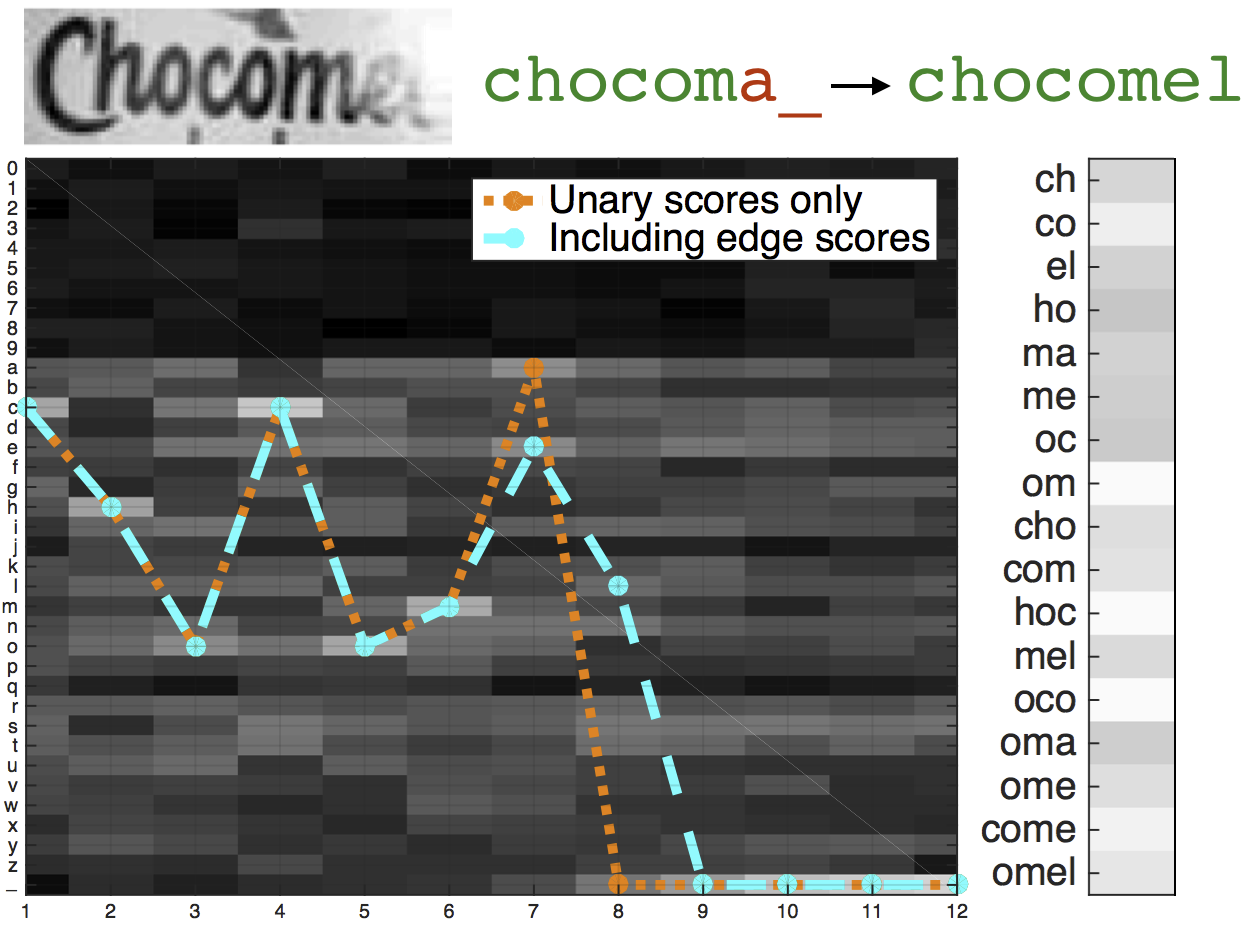}\\
(a) & (b) & (c)
\end{tabular}
\vspace{-1.4em}
\caption{\small Results where the unary terms of the JOINT model cannot solely recognise the word correctly, but the addition of the edge scores result in correct recognition, from SVT (a,b) and IC13 (c). The input image is shown in the top left corner. The unary scores for characters (rows) at each position (columns, up to 12 out of 23 characters) are shown, with the selected path using only the unary score term $S^i_c$ (orange) and when edge scores $S_e$ are incorporated (cyan). The bars show the NGRAM strengths, with lighter colour representing larger values.}
\label{fig:results}
\end{center}
\end{figure}

The CHAR and JOINT models are evaluated on standard as well as synthetic benchmarks (Table~\ref{table:internal}), but both models are trained on Synth90k. While the CHAR model achieves good performance, it is consistently outperformed by the JOINT model; the accuracy improvement is as much as +4\% on IC03 and SVT, despite the difficulty of the latter. Figure~\ref{fig:results} shows some example results using the JOINT model.

Next, we evaluate the ability of our model to generalise by
recognising words unseen during training. This effectively amounts to zero-shot learning and is a key contribution compared to~\cite{Jaderberg14c,Jaderberg14d}.
In order to do so, the training vocabulary is split into two parts, with one part (50\% or 80\%) used for training and the other one for evaluation (50\% or 20\%). In this case the CHAR model is significantly penalised, but the JOINT model can recover most of the performance. For instance, on the 50/50 split, the JOINT model accuracy is 89.1\%, only -2\% compared to the 91.0\% obtained when the training and testing vocabularies are equal.

The final test pushes generalisation by training and testing on completely random strings from SynthRand. As this dataset is a lot less regular than a natural language, the performance of the CHAR model suffers, dropping to 80.7\% accuracy. Furthermore, as could be expected form the absence of common N-grams in the random language, the JOINT model performs \emph{slightly worse} at 79.5\% accuracy. However this drop is very small because N-grams are not used as hard constraints on the predicted words, but rather to nudge the word scores based on further visual cues.

\paragraph{Comparison to the state-of-the-art.}

Table~\ref{table:comparison} compares the accuracy of CHAR and JOINT to previous works. Whereas these works make use of strong language models, our models make minimal assumptions about the language. In the constrained lexicon cases (the starred columns of Table~\ref{table:comparison}), both CHAR and JOINT are very close to the state-of-the-art DICT model of \cite{Jaderberg14c,Jaderberg14d}. Furthermore, if the same 90k dictionary used in by the DICT model is used to constrain the output of the JOINT model, the performance is identical at 93.1\% accuracy on IC03. While in the constrained lexicon experiments the lexicon is limited at test time, these results are still remarkable because, differently from DICT, CHAR and JOINT are not trained on a specific dictionary. In particular, DICT would not be able to operate on random strings.

The recognition results without a lexicon are still behind that of some constrained models, however the JOINT model provides competitive performance and is far more flexible to recognise unseen words than previous works, while still achieving state-of-the-art performance if a lexicon is then applied as a constraint at test time. Figure~\ref{fig:results} shows some example results where the CHAR model does not recognise the word correctly but the JOINT model succeeds. 

\setlength{\tabcolsep}{3pt}
\begin{table}[t]
\begin{center}
\scriptsize
\begin{tabular}[t]{|l||c|c|c|c|c|c|c|c|c|} 
\hline
\multicolumn{1}{|c||}{\centering Model} & 
\multicolumn{1}{c|}{\centering IC03-50\textsuperscript{*}} &
\multicolumn{1}{c|}{\centering IC03-Full\textsuperscript{*}} &
\multicolumn{1}{c|}{\centering IC03-50k\textsuperscript{*}} &
\multicolumn{1}{c|}{\centering IC03} &
\multicolumn{1}{c|}{\centering SVT-50\textsuperscript{*}} &
\multicolumn{1}{c|}{\centering SVT} &
\multicolumn{1}{c|}{\centering IC13} &
\multicolumn{1}{c|}{\centering IIIT5k-50\textsuperscript{*}} &
\multicolumn{1}{c|}{\centering IIIT5k-1k\textsuperscript{*}} \\
\hline\hline
\rowcolor{Gray}
\emph{Baseline ABBYY} (\cite{Wang11}) & 56.0 & 55.0 & - & - & 35.0 & - & - & 24.3 & -\\
\cite{Wang11}          & 76.0 & 62.0 & - & - & 57.0 & - & - & - & -\\
\rowcolor{Gray}
\cite{Mishra12}      & 81.8 & 67.8 & - & - & 73.2 & - & - & - & -\\
\cite{Novikova12}  & 82.8 & - & - & - & 72.9 & - & - & 64.1 & 57.5\\
\rowcolor{Gray}
\cite{Wang12}    & 90.0 & 84.0 & - & - & 70.0 & - & - & - & -\\
\cite{Goel13}          & 89.7 & - & - & - & 77.3 & - & - & - & -\\
\rowcolor{Gray}
\cite{Bissacco13}& - & - & - & - & 90.4 & 78.0 & 87.6 & - & -\\
\cite{Alsharif13}  & 93.1 & 88.6 & 85.1 & - & 74.3 & - & - & - & -\\
\rowcolor{Gray}
\cite{Almazan14} & - & -  & -& - & 89.2 & - & - & 91.2 & 82.1\\
\cite{Yao14}  & 88.5 & 80.3 & - & - & 75.9 & - & - & 80.2 & 69.3\\
\rowcolor{Gray}
\cite{Jaderberg14a}  & 96.2 & 91.5 & - & - & 86.1 & - & - & - & -\\
\cite{Gordo14} & - & - & - & - & 90.7 & - & - & 93.3 & 86.6\\
\rowcolor{Gray}
DICT \cite{Jaderberg14c,Jaderberg14d} & \bf 98.7 & \bf 98.6 &  93.3 & \bf 93.1 & \bf 95.4 & \bf 80.7 & \bf 90.8 & \bf 97.1 & \bf 92.7\\
\hline
CHAR     &  98.5 &  96.7 &  92.3 & 85.9 &  93.5 & 68.0 & 79.5 &  95.0 & 89.3\\

\rowcolor{Gray}
JOINT     &  97.8 &  97.0 &  \bf 93.4 & 89.6 &  93.2 & 71.7 & 81.8 & 95.5 & 89.6\\

\hline
\end{tabular}
\end{center}
\vspace*{-1em}
\caption{\small Comparison to previous methods. The baseline method is from a commercially available OCR system. Note that the training data for DICT includes the lexicons of the test sets, so it has the capacity to recognise all test words. \textsuperscript{*}Results are constrained to the lexicons described in Section~\ref{sec:datasets}.}
\label{table:comparison}
\end{table}

\section{Conclusion}
\label{sec:conclusion}
In this paper we have introduced a new formulation for word
recognition, designed to be used identically in language and
non-language scenarios. By modelling character positions and the
presence of common N-grams, we can define a joint graphical
model. This can be trained effectively by back propagating structured
output loss, and results in a more accurate word recognition
system than predicting characters alone. We show impressive results for unconstrained text recognition
with the ability to generalise recognition to previously unseen words,
and match state-of-the-art accuracy when comparing in lexicon
constrained scenarios. 


\paragraph{Acknowledgments.}

This work was supported by the EPSRC and ERC grant VisRec no. 228180. We gratefully acknowledge the support of NVIDIA Corporation with the donation of the GPUs used for this research.

\bibliographystyle{iclr2015}
{
   \small
   \setlength{\bibsep}{4pt}
   \bibliography{shortstrings,vgg_local,vgg_other,current,max_bib}
}

\end{document}